\documentclass{article}
\usepackage{graphicx}

\usepackage{multirow}
\usepackage[makeroom]{cancel}
\usepackage{amsmath}
\usepackage{xcolor}
\usepackage[margin=4.28cm]{geometry}
\begin{document}
\title{DETECT: A Hierarchical Clustering Algorithm for Behavioural Trends in Temporal Educational Data}
\author{Jessica McBroom, Kalina Yacef and Irena Koprinska \\ School of Computer Science, University of Sydney, Sydney, Australia \\ \{jmcb6755, kalina.yacef, irena.koprinska\}@sydney.edu.au}
\date{May 4, 2020}

\maketitle

\begin{abstract}
Techniques for clustering student behaviour offer many opportunities to improve educational outcomes by providing insight into student learning. However, one important aspect of student behaviour, namely its evolution over time, can often be challenging to identify using existing methods. This is because the objective functions used by these methods do not explicitly aim to find cluster trends in time, so these trends may not be clearly represented in the results. This paper presents `DETECT' (\textit{\textbf{D}etection of \textbf{E}ducational \textbf{T}rends \textbf{E}licited by \textbf{C}lustering \textbf{T}ime-series data}), a novel divisive hierarchical clustering algorithm that incorporates temporal information into its objective function to prioritise the detection of behavioural trends. The resulting clusters are similar in structure to a decision tree, with a hierarchy of clusters defined by decision rules on features. DETECT is easy to apply, highly customisable, applicable to a wide range of educational datasets and yields easily interpretable results. Through a case study of two online programming courses ($N>600$), this paper demonstrates two example applications of DETECT: 1) to identify how cohort behaviour develops over time and 2) to identify student behaviours that characterise exercises where many students give up.

\end{abstract}

\section{Introduction}

In recent decades, educational datasets have become increasingly rich and complex, offering many opportunities for analysing student behaviour to improve educational outcomes. The analysis of student behaviour, particularly temporal trends in this behaviour, has played a major role in many recent studies in areas including automated feedback provision \cite{keuning2018systematic,mcbroom2019survey,paassen2016adaptive,price2019comparison}, dropout analysis \cite{pal2012mining,pereira2019early,tang2018time}, collaborative learning \cite{barros2000analysing,talavera2004mining,villamor2017assessing} and student equity \cite{finkelstein2013effects,makhija2018influence,mcbroom2020understanding}.

However, a significant challenge in analysing student behaviour is its complexity and diversity. As such, clustering techniques \cite{xu2015comprehensive}, which organise complex data into simpler subsets, are an important resource for analysing student behaviour, and have been employed in many recent studies \cite{dutt2015clustering}. For example,  in \cite{adjei2017clustering} and \cite{alias2017mining} K-means clustering and a self-organising map, respectively, are used to group students based on their interactions with an educational system. In addition, in \cite{joyner2019clusters} student programs are clustered to identify common misconceptions.

One limitation of standard clustering techniques is that they are not well-suited to detecting behavioural trends in time. One solution is to use time-series clustering techniques, which typically combine standard techniques with extra processing steps \cite{aghabozorgi2015time}. For example, \cite{mlynarska2016time} uses dynamic time warping in conjunction with K-means clustering to cluster time series' of student Moodle activity data. Alternatively, temporal information is often considered only after all student work samples or behaviours have been clustered. For example, in \cite{mcbroom2018data} student work is clustered to allow an interaction network over time to be built and in \cite{mcbroom2016mining}  clusters of student behaviour over time are used in a second round of clustering.

Although it is possible to gain insight into student behavioural changes using these techniques, one important limitation is that temporal trend detection is not explicitly incorporated into the objective function when clustering. For example, consider the case where K-means is first used to cluster student behaviours, and then cluster changes over time are observed, as in \cite{mcbroom2016mining}. Since the objective of K-means is to minimise the distance between points (which in this case represent student behaviours), the process will prioritise grouping behaviours that match on as many features as possible. However, this may obscure important trends, especially if many of the features are unrelated to these trends.

The contribution of this paper is `DETECT' (\textit{\textbf{D}etection of \textbf{E}ducational \textbf{T}rends \textbf{E}licited by \textbf{C}lustering \textbf{T}ime-series data}), a novel divisive hierarchical clustering algorithm that incorporates trend detection into its objective function in order to identify interesting patterns in student behaviour over time. DETECT is highly general and can be applied to many educational datasets with temporal data (for example, from regular homework tasks or repeated activities). In addition, it can be customised to detect a variety of trends and produces clusters that are well-defined and easy to understand. Moreover, it does not require that the features be independent, or that the objective function be differentiable.

Broadly, DETECT has similar properties to the classification technique of decision trees \cite{safavian1991survey}. In particular, it produces a hierarchy of clusters distinguished by decision rules. However, whereas decision trees are a supervised technique requiring the existence of classes in order to calculate entropy, DETECT is unsupervised and uses an objective function completely unrelated to this measure.

This paper is set out as follows: Section \ref{sec:clustering_algorithm} describes the DETECT algorithm, including the input it takes, its flexibility and how the output is interpreted. Section \ref{sec:case_study} then shows example usage of the algorithm through a case study and Section \ref{sec:conclusion} concludes with a summary of the main ideas of the paper. 

\section{DETECT Algorithm} \label{sec:clustering_algorithm}

\subsection{Overview} \label{sec:overview}

DETECT produces clusters of student behaviour that reveal cohort behavioural trends in educational datasets. Such trends can include changes in behaviour over time, anomalous behaviours at specific points in the course or a variety of other customisable trends. This is achieved by iteratively dividing student behaviours into clusters that maximise a time-based objective function. The clusters found can then be interpreted by teachers and course designers to better understand student behaviour during the course.

DETECT can be applied to a wide range of datasets, of the form described in Table \ref{tab1}. In particular, the data should be \textit{temporal} - that is, able to be divided into a series of comparable time steps. For example, a series of homework tasks or fixed time periods during an intervention could be considered as comparable time steps. In addition, for each student and time period, there should be a set of features describing the behaviour of the student during that time period. These features could be numeric, such as the number of exercise attempts, or categorical, such as a label for the style of their work. Note that features are not required to be independent or equally important, since the objective function can determine the quality of features and penalise less useful ones.

\begin{table}
	\centering
	\caption{Structure of input data, where the number of cells in the table is equal to S (number of students) $\times$ T (number of time periods) $\times$ M (number of features).  F1,...,FM are different feature names.}\label{tab1}
	\begin{tabular}{|l|l|l|l|l|l|}
		\hline
		\textbf{Student} &  \textbf{Time} & \textbf{F1} & \textbf{F2} & ... & \textbf{FM}\\
		\hline
		1 &  1 & $v_{111}$ & $v_{112}$ & ... & $v_{11M}$\\
		1 &  2 & $v_{121}$ & $v_{122}$ & ... & $v_{12M}$\\
		... &  ... & ... & ... & ... &...\\
		1 &  T & $v_{1T1}$ & $v_{1T2}$ & ... & $v_{1TM}$\\
		... &  ... & ... & ... & ... &...\\
		S &  1 & $v_{N11}$ & $v_{N12}$ & ... & $v_{N1M}$\\
		... &  ... & ... & ... & ... &...\\
		S &  T & $v_{NT1}$ & $v_{NT2}$ & ... & $v_{NTM}$\\
		\hline
	\end{tabular}
\end{table}

DETECT outputs clusters of student behaviour explicitly defined by rules on feature values. For example, a cluster may be defined as all rows of the input data where $\textsf{`}num\_submissions\textsf{'} \leq 7$ and  $\textsf{`}completed\textsf{'} == \textsf{`}yes\textsf{'}$. These clusters are organised into a hierarchical structure where, in each successive level, an additional condition is added, similarly to a decision tree.  Examples of this are given in Section \ref{sec:case_study}, as part of the case study. 

It is important to note that the clusters are not clusters of students but rather clusters of the input data rows (which each represent the behaviour of one student at one time period).  This means students are in many clusters - one for each time period. By observing changes to the distributions of clusters over time, trends in student behaviour can be identified (see Section \ref{sec:case_study}).

\subsection{Cluster Formation}

\begin{figure}[h!]
	\centering
	\includegraphics[width=0.9\textwidth]{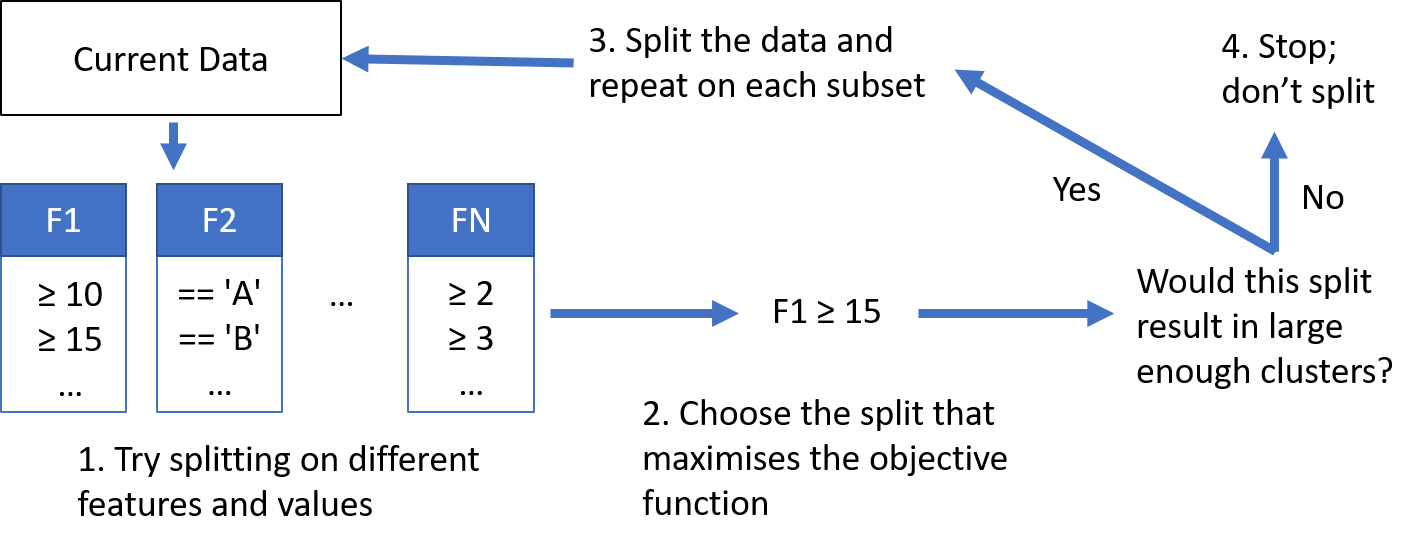}
	\caption{A summary of the steps involved in DETECT.} \label{fig1}
\end{figure}

Clusters are formed divisively through an iterative process with four main steps, as summarised in Figure \ref{fig1}. Initially, all examples are placed in the same cluster. Then, during each iteration, a search is performed to find the best feature and value to split this cluster on. If this split would result in new clusters that are larger in size than a specified lower-limit (e.g. at least 100 examples each), then the split it performed, creating two new clusters, and the process is repeated recursively on the new clusters. Otherwise, the split is not performed. The algorithm terminates when no cluster can be split further.

Using the given objective functions and assuming the cluster size threshold scales proportionally with the number of examples (which places a constant upper bound on the number of clusters), the time complexity of this process is $O(nm\log n)$, where $m$ is the number of features and $n$ the number of examples.

\subsubsection{Feature and Value Search.} Before a cluster is divided, a search is performed to find the best feature and value to split on. For each feature, this can be performed in $O(n\log n)$ time (where n is the cluster size) using the given objective functions in the next subsection. For numeric features, the process is as follows:

\begin{enumerate}
	\item Sort the examples in ascending order based on the feature value.
	\item Create two new clusters, one containing no examples ($C_a$) and the other containing all ($C_b$).
	\item Set a threshold, $t$, that is lower than the smallest feature value.
	\item While there are still examples in $C_b$, increase $t$ to the next smallest feature value (or larger) and add all examples $ \leq t$ from $C_b$ to $C_a$, each time checking if this improves the objective function value (and, if so, remembering $t$).
\end{enumerate}

The best feature and split will then be the one that optimises the function values. Similarly, for categorical features, each category can be iterated through to find the best one to split off from the rest. Note that we recommend minimising the amount of missing data (e.g. by selecting subsets of students or making time periods relative to students as in \cite{mcbroom2020how}). However, if required, missing values can be treated as another category if the feature is categorical, or, if the feature is numeric, the process can be repeated twice -- once where the missing value examples are always in $C_a$, and once where they are always in $C_b$.

\subsubsection{Objective Function.} \label{sec:opt_func} 
The objective function is used to determine the quality of potential cluster divisions using temporal information, thereby controlling the types of trends detected by the algorithm. More specifically, this function maps the distributions of $C_a$ and $C_b$ over time, along with optional additional parameters, to a score. It can be customised to suit different purposes and there are no constraints such as differentiability on the function. Two examples of an objective function are defined here:

\

\noindent Let $\textbf{n} = [n_a(t_i),n_b(t_i)]$ be the number of students in clusters $C_a$ and $C_b$ respectively at time $i$ and $T$ be the number of time steps in total. In addition, for $f_2$, let $x$ be a time step of interest. Then:
\begin{align*}
f_1(\textbf{n}(t_1),\textbf{n}(t_2),...,\textbf{n}(t_T)) &= \left|\frac{n_a(t_1) + n_a(t_2)}{2} - \frac{n_a(t_T) + n_a(t_{T-1})}{2}\right| \\
f_2(\textbf{n}(t_1),\textbf{n}(t_2),...,\textbf{n}(t_T), x) &= \frac{\left|n_a(t_x) - n_a(t_{x+1})\right| + \left|n_a(t_x) - n_a(t_{x-1})\right|}{2}
\end{align*}

The first function, $f_1$, compares the average number of students in $C_a$ at the beginning of the course to the average at the end. As such, it is a measure of how many students change cluster from the start to the end of the course, and will be maximised when there is a large shift in behaviour. In contrast, the second function, $f_2$, compares the number of students in $C_a$ at time $x$ to the adjacent time periods and finds the average difference. As such, it is maximised when behaviours at time $x$ vary greatly from those at neighbouring times.

\section{Case Study} \label{sec:case_study}
This section demonstrates two example applications of DETECT using the two objective functions introduced in Sec. \ref{sec:opt_func}. Specifically, in the first example we apply DETECT to an intermediate course using $f_1$ to find behavioural trends over time. In the second example, we then apply DETECT to a beginner course using $f_2$ to find behaviours that characterise an exercise where many students give up. While the data come from programming courses, we only use general features not specific to this domain to demonstrate the generality of the approach.

\subsection{Data} \label{sec:data}
Our data come from school students participating in two online Python programming courses of different difficulty levels: intermediate (N=4213) and beginners (N=7164).\footnote{\label{note1} These refer to the number of students who attempted at least the first exercise.} These courses were held over a 5-week period during 2018 as part of a programming challenge held primarily in Australia. The courses involved weekly notes, which introduced students to new concepts, and programming exercises to practice these skills. Students received automated feedback on their work from test cases and were able to improve and resubmit their work.

From this data, we extracted 10 features per student per exercise: 1-3) the number of times the student viewed, failed and passed the exercise, 4) the number of times their work was automatically saved (triggered when unsaved work was left for 10 seconds without being edited), 5-8) the time of the first view, autosave, fail and pass relative to the deadline, 9) the average time between successive fails and 10) the time between the first fail and passing. Note that these features did not need to be independent (see Section \ref{sec:overview}).

\subsection{Example 1: Using $f_1$ to Detect Changes Over Time}
\begin{figure}[h!]
	\centering
	\includegraphics[width=0.86\textwidth]{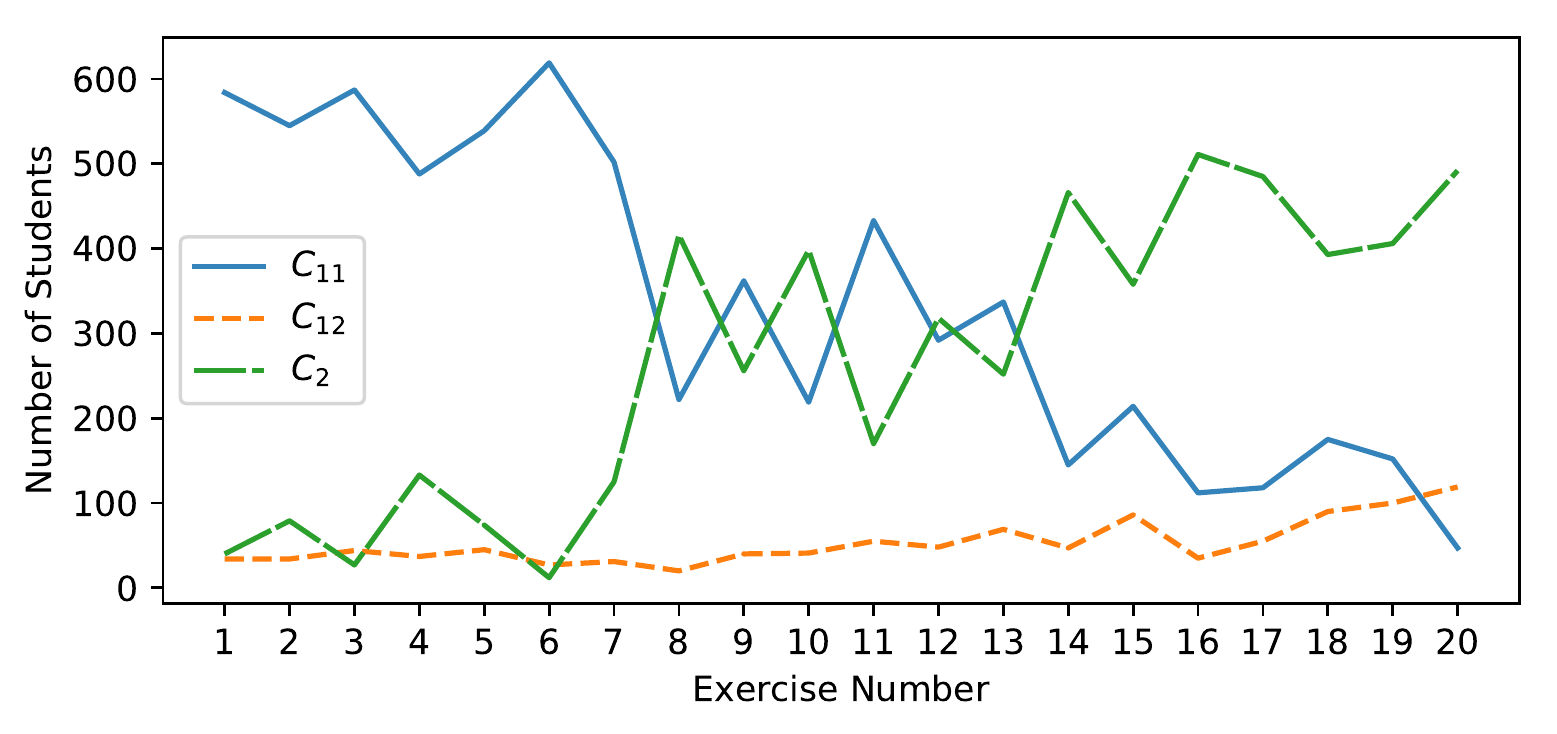}
	\caption{The distribution of final clusters over time when $f_1$ is applied to data from all students who completed the course. Most students begin in $C_{11}$, but transition to other clusters over time.} \label{fig_change_completing}
\end{figure}

When analysing student behaviour during a course, one important question is how this behaviour changes in time. To answer this, $f_1$ was applied to the data from the intermediate course. Since exercises from the last week differed in structure from the others (i.e. students were given significantly less time to complete them), these were excluded, leaving a total of 20 exercises. Each of the remaining exercises were then considered as a time period. The resulting behavioural clusters are given in Table \ref{tab5} and the number of students in each of the final clusters at each time period is shown in Figure \ref{fig_change_completing}. Note that only data from completing students (N=658) was used to minimise the amount of missing data.

\begin{table}
	\centering
	\caption{Clusters formed by applying $f_1$ to the intermediate course, using a minimum cluster size threshold of 400 - i.e. an average of 20 students per time period. The clusters are defined by the number of autosaves (level 1) and how long before the deadline the exercise was completed (level 2).}
	\label{tab5}
	\begin{tabular}{l l l}
		\hline
		\textbf{Level 1} &  \textbf{Level 2} & \textbf{Label} \\
		\hline
		\multirow{2}{*}{autosaves $\leq$ 9}   
		& completed 7.25 days or more before deadline  & $C_{11}$  \\ \cline{2-3}
		& completed within 7.25 days of deadline  & $C_{12}$\\
		\hline
		autosaves $>$ 9 &   & $C_{2}$  \\
		\hline
	\end{tabular}
\end{table}

From Figure \ref{fig_change_completing} and Table \ref{tab5}, the most important difference in behaviour between the beginning and end of the course was the number of autosaves, which increased over time. In particular, Figure \ref{fig_change_completing} shows that most students began in $C_{11}$ (with $\leq 9$ autosaves) and ended in $C_{2}$ (with $> 9$ autosaves). Since these autosaves were triggered when a student paused for 10 or more seconds, this could indicate increased difficulty (if the students were pausing to read notes or think) or increased disengagement (if they were frequently stopping to do other activities). 

Interestingly, even the students who had a smaller number of autosaves changed in behaviour over time, with an increasing proportion completing the exercises closer to the deadline as the course progressed. This can be seen by the increasing proportion of students in $C_{12}$ compared to $C_{11}$ over time. For the importance of this change to be apparent, note that passing these exercises within 7 days of the deadline actually indicates that a student is behind schedule. This is because each week of exercises is intended to take one week, but the deadlines for the first four weeks allow two weeks. If students were falling behind over time, this may suggest that the course content was too dense, and perhaps reducing the amount of content or spreading the course over a larger time period could be beneficial for students.

Furthermore, the distribution of clusters does not change smoothly over time. In particular, the plot lines are jagged, indicating that student behaviour varies a lot even between adjacent exercises. This is particularly interesting considering the features the clusters are defined by. For example, the fact that the number of autosaves varies a lot between adjacent exercises indicates that some exercises may be more interesting or difficult than similar exercises. For instance, the number of students in $C_{11}$ (where there are $\leq 9$ autosaves) drops by almost 100 from Exercise 3 to Exercise 4, then increases again at Exercise 5, even though all three exercises involve if-else statements. This could indicate that Exercise 4 is more difficult or less interesting than the others, since students pause more here (either because they are thinking or doing something else).

Another interesting observation is that there are three general and overarching changes to the cluster distributions over the course. In particular, from around Exercises 1 to 7, $C_{11}$ is most dominant. Upon inspection, these exercises are primarily revision exercises (e.g. printing, variables and if-statements). After this period, there is an immediate shift in cluster distributions, with $C_{11}$ and $C_{2}$ becoming similar in size, as students begin to learn about string slicing and loops. This general change suggests that students may find these topics more challenging than the previous ones. After Exercise 13, $C_2$ becomes dominant and $C_{12}$ overtakes $C_{11}$ in size as students learn about list operations, dictionaries and files. Since these general changes in student behaviour seem to occur as the topics become increasingly complex, perhaps the course could be improved by condensing the large revision period and expanding the other topics to allow for a more gradual difficulty change.

In summary, even by using DETECT with a simple objective function, $f_1$, and a highly general set of features, distinct and interpretable clusters can be found that coherently represent changes in student behaviour over time. By observing how the distributions of these clusters change at different scales (i.e. over the whole course, over groups of exercises or between individual exercises), important insights into student behaviour can be easily gained, and then used for purposes such as informing course development.

\subsection{Example 2: Using $f_2$ to Analyse Behaviour Where Many Students Quit}

Another topic of interest when analysing a course is the exercises that students have difficulty completing. In particular, if students attempt an exercise but cannot complete it, this can discourage them from continuing and lead to increased disengagement. This is particularly concerning in a beginner course, where students may not yet have confidence and could be dissuaded from pursuing further study in the area. This section provides an example of how DETECT could be used with objective function $f_2$ to explore such issues.

\begin{figure}[h!]
	\centering
	\includegraphics[width=0.86\textwidth]{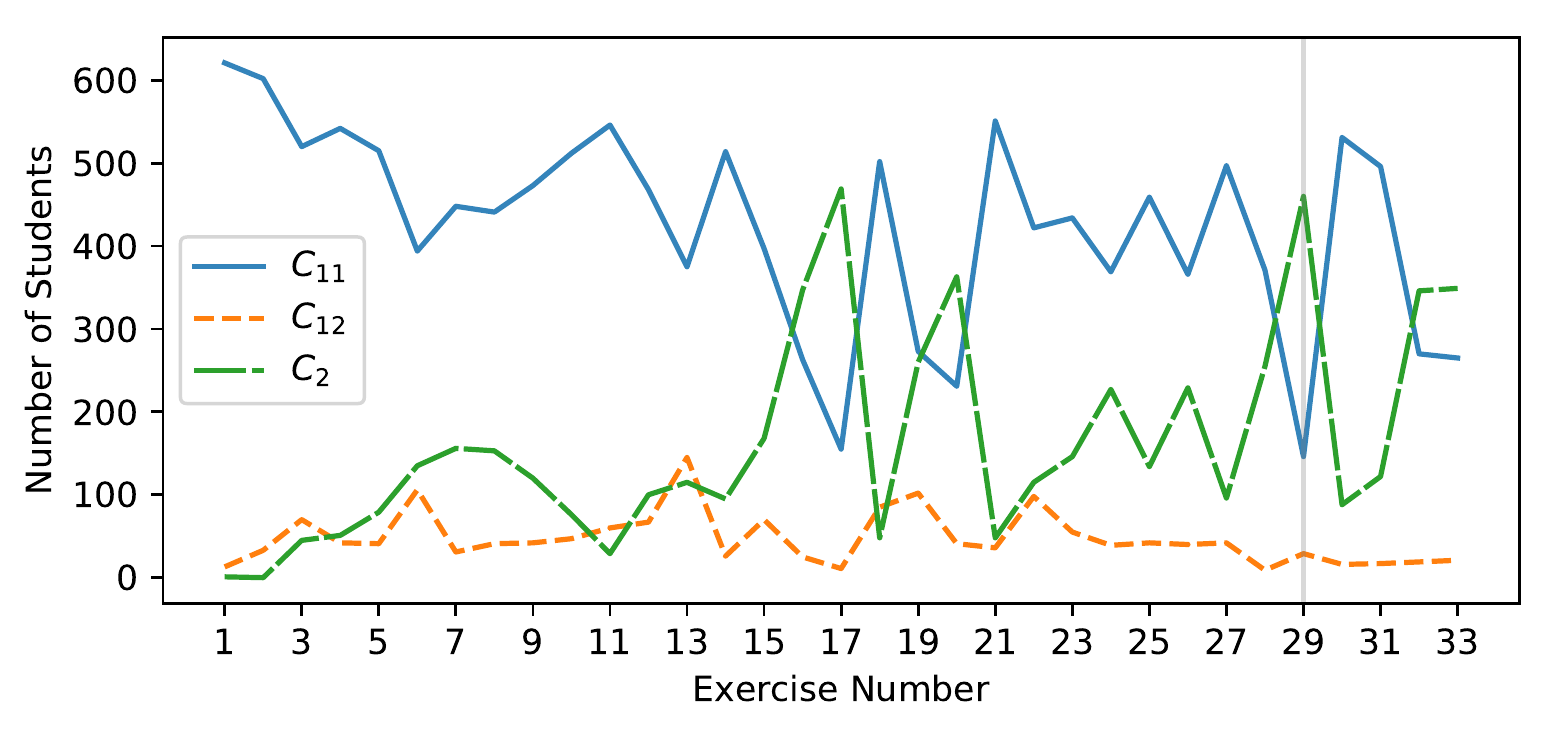}
	\caption{The distribution of final clusters over time when $f_2$ is applied to data from all students who completed the course using Exercise 29 (marked in grey) as a parameter.} \label{fig_peak}
	
	\includegraphics[width=0.86\textwidth]{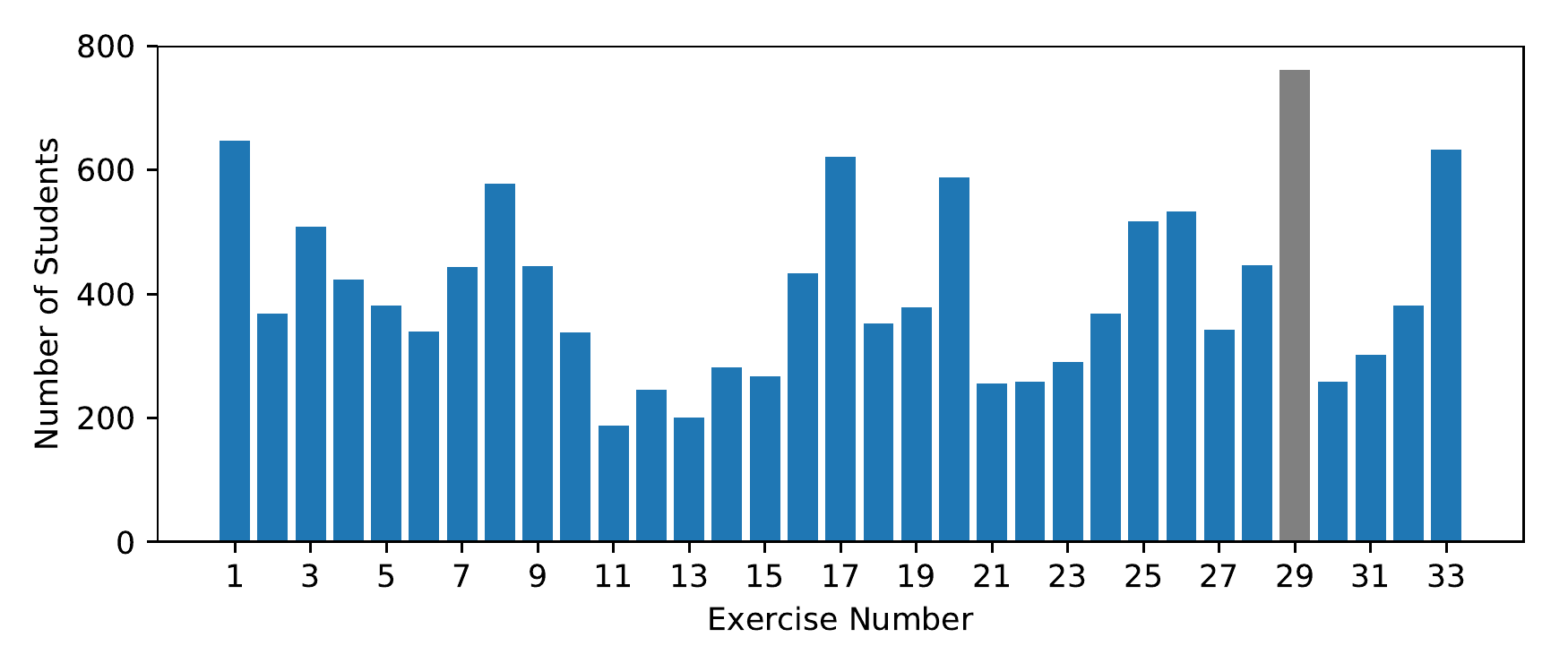}
	\caption{The number of students who attempt but do not complete each of the beginner exercises from the first four weeks. Exercise 29, used for clustering, is marked in grey.} \label{fig_failures}
\end{figure}

During the beginner course, 761 students attempted but could not complete Exercise 29 - the highest out of any exercise during the first four weeks. To understand how student behaviour differed during this exercise compared to others, we applied DETECT to the data using $f_2$ (setting $x=29$), which identified clusters that distinguished this exercise from adjacent ones. The clusters formed and their distributions over time are shown in Table \ref{tab6} and Figure \ref{fig_peak} respectively.

\begin{table}
	\centering
	\caption{Clusters formed by applying $f_2$ to data from completing beginner students ($N=635$) with parameter $x=29$ and minimum size threshold of 660 (i.e. an average of 20 students per time period). The clusters are defined by the number of autosaves and the time between a student's first failure and completion of the exercise.}
	\label{tab6}
	\begin{tabular}{lp{7.8cm}l}
		\hline
		\textbf{Level 1} &  \textbf{Level 2} & \textbf{Label} \\
		\hline
		\multirow{2}{*}{autosaves $\leq$ 8}   
		&  time from first fail to completion $\leq$ 48 secs, or no fails  & $C_{11}$ \\
		\cline{2-3}
		&   time from first fail to completion $>$ 48 secs  & $C_{12}$\\ \hline
		autosaves $>$ 8 &   & $C_{2}$  \\
		\hline
	\end{tabular}
\end{table}

By comparing the cluster distribution at Exercise 29 to the adjacent exercises in Figure \ref{fig_peak}, three general differences can be observed. Firstly, the proportion of students in $C_2$ ($> 8$ autosaves) is much higher for Exercise 29, indicating that students paused more often. In addition, the proportion of students in $C_{11}$ is much lower. Since this cluster describes behaviour where students quickly solve the task (i.e. with few pauses, and either no failed submissions or a short time from their first fail to passing), a decrease in its frequency suggests this exercise is especially challenging compared to the adjacent tasks. The slight increase in the frequency of $C_{12}$ (where the time from first failing to passing is $> 48$ secs) supports this, suggesting students take longer to correct their work after failing.

Interestingly, the pattern of $C_2$ sharply increasing and $C_{11}$ sharply decreasing is not limited to Exercise 29. For example, this change also occurs at Exercises 17 and 20. From Figure \ref{fig_failures}, which shows the number of students who unsuccessfully attempted each exercise, Exercises 17 and 20 also appear to have resulted in a large number of students giving up, especially relative to the adjacent exercises. Since information about these exercises was not used in the clustering, this is a strong indication that the cluster changes are not simply noise, but meaningful behaviour associated with times when students give up.\footnote{Indeed, regression analysis finds that the correlation between the percentage of students in $C_2$ and the number of students unsuccessfully attempting each exercise is statistically significant with a p-value of 0.008.} 

Since the clusters are distinguished by the number of autosaves and also the time between a student's first fail and completion, one potential use of this information could be to improve interventions. For example, additional feedback or support messages could be triggered if a student pauses too many times or is unsuccessful in correcting their work for too long after their first fail. In addition, since students already receive automated feedback after failing an exercise, perhaps longer correction times could indicate that this feedback is unclear and could be revised. Finally, perhaps the information could be a useful tool when testing future courses. For example, senior students or a teacher could test-run a course, and the relative differences in the number of autosaves or time taken after failing could be used to highlight potential issues in advance.

In summary, this example has demonstrated how DETECT can be used to find different kinds of trends in educational data by changing the objective function. This customisable feature allows for great flexibility so that DETECT can be used for a range of interesting purposes.

\section{Conclusion} \label{sec:conclusion}
This paper has presented a novel hierarchical clustering algorithm, DETECT, for identifying behavioural trends in temporal educational data. In contrast to current clustering approaches, DETECT incorporates temporal information into its objective function to prioritise the detection of behavioural trends. It can be applied to a wide range of educational datasets, produces easily interpretable results and is easy to apply, since the input features do not need to be independent. Two examples of objective functions have been provided, but these can be customised to identify different trends with few constraints (e.g. the functions do not need to be differentiable). 

Through a case study, this paper has shown how DETECT can be used to identify interesting behavioural trends in educational data, even when the features are simple and not domain-specific. In particular, it can detect general changes in student behaviour over time or highlight behaviours characterising exercises where students give up. Such information is invaluable to teachers, course designers and researchers, who can use it to understand student behaviour, stimulate further investigation and ultimately improve educational outcomes.

In future, we hope to further develop DETECT by considering a greater range of objective functions and stopping conditions, and exploring the impact of additional domain-specific features and missing data on trend detection. In addition, it would be interesting to consider how DETECT could be used in conjunction with other techniques to, for example, analyse individual student trajectories. Ultimately, in a time when educational data are becoming increasingly abundant, this work aims to contribute to better-understanding student behaviour in order to improve educational outcomes.

\bibliographystyle{plain}
\bibliography{references}

\end{document}